\documentclass[11pt,a4paper]{article}
\usepackage{times,latexsym}
\usepackage{url}
\usepackage[T1]{fontenc}
\usepackage{comment}
\usepackage{xfrac}
\usepackage[font=small,labelfont=bf]{caption}

\usepackage[acceptedWithA]{tacl2018v2}

\usepackage{times}
\usepackage{amsmath}
\usepackage{amssymb}
\usepackage{amsfonts}
\usepackage{bbm}
\usepackage{color}
\usepackage{latexsym}
\usepackage{booktabs}
\usepackage{fontawesome}

\usepackage{cleveref}
\crefname{section}{\S}{\S\S}
\Crefname{section}{\S}{\S\S}
\crefname{table}{Tab.}{}
\crefname{figure}{Fig.}{}
\crefname{algorithm}{Alg.}{}
\crefname{appendix}{App.}{}
\crefname{lemma}{Lemma}{}
\Crefname{theorem}{Theorem}{}
\crefname{prop}{Proposition}{}
\crefname{cor}{Corollary}{}
\crefname{align}{}{}
\crefname{equation}{}{}

\usepackage{xcolor}
\definecolor{myorange}{RGB}{245, 121, 58}
\definecolor{mypurple}{RGB}{169, 90, 161}

\usepackage{algorithm}
\usepackage{algorithmicx}
\usepackage[noend]{algpseudocode}
\newcommand{\rightcomment}[1]{{ \(\triangleright\) {\footnotesize\textit{#1}}}}
\algrenewcommand{\algorithmiccomment}[1]{\hfill \rightcomment{#1}}  %
\algnewcommand{\LineComment}[1]{\State \rightcomment{#1}}
\algnewcommand{\LinesComment}[1]{\State \rightcomment{\parbox[t]{\linewidth-\leftmargin-\widthof{\(\triangleright\) }}{#1}}}

\usepackage{textcomp}
\usepackage{subcaption}
\usepackage{graphicx}

\usepackage{microtype}
\usepackage[disable]{todonotes}

\usepackage{float}
\everypar{\looseness=-1}

\newcommand{\softmax}{\mathrm{softmax}}

\newcommand{\att}{\mathrm{att}}

\newcommand{\mhatt}{\mathrm{mhatt}}
\newcommand{\gmhatt}{\mathrm{gmhatt}}

\newcommand{\calD}{\mathcal{D}}
\newcommand{\bert}{\small \textsf{BERT}}
\newcommand{\ed}{\small \textsf{Enc--Dec}}
\newcommand{\defn}[1]{\textbf{#1}}
\newcommand{\R}{\mathbb{R}}
\newcommand{\xx}{\mathbf{x}}

\newcommand{\zz}{\mathbf{z}}

\newcommand{\yy}{\mathbf{y}}
\newcommand{\calL}{\mathcal{L}}
\newcommand{\calH}{\mathcal{H}}
\newcommand{\calJ}{\mathcal{J}}

\newcommand{\calR}{\mathcal{R}}

\DeclareMathOperator*{\argmax}{argmax}

\usepackage{fdsymbol}
\newcommand{\ethzsymbol}{\clubsuit}
\newcommand{\cambsymbol}{\spadesuit}
\newcommand{\ethz}{$^\ethzsymbol$}
\newcommand{\camb}{$^\cambsymbol$}

\title{Differentiable Subset Pruning of Transformer Heads}

\author{
 Jiaoda Li\ethz \quad Ryan Cotterell{\ethz\camb} \quad Mrinmaya Sachan\ethz \\
 \ethz ETH Z\"{u}rich \quad \camb University of Cambridge\\
  {\sf \texttt{\{jiaoda.li,ryan.cotterell,mrinmaya.sachan\}@inf.ethz.ch}} \\
}

\date{}

\begin{document}
\maketitle
\begin{abstract}
Multi-head attention, a collection of several attention mechanisms that independently attend to different parts of the input, is the key ingredient in the Transformer.
Recent work has shown, however, that a large proportion of the heads in a Transformer's multi-head attention mechanism can be safely pruned away without significantly harming the performance of the model; such pruning leads to models that are noticeably smaller and faster in practice.
Our work introduces a new head pruning technique that we term differentiable subset pruning.
Intuitively, our method learns per-head importance variables and then enforces a user-specified hard constraint on the number of unpruned heads.
The importance variables are learned via stochastic gradient descent. 
We conduct experiments on natural language inference and machine translation; we show that differentiable subset pruning performs comparably or better than previous works while offering precise control of the sparsity level.\looseness=-1\footnote{\scriptsize \texttt{\url{https://github.com/rycolab/differentiable-subset-pruning}.}}
\end{abstract}

\section{Introduction}\label{sec:introduction}
The Transformer \cite{vaswani2017attention}
has become one of the most popular neural architectures used in NLP. 
Adaptations of the Transformer have been applied to nearly every popular NLP task, e.g. parsing \cite{zhou-zhao-2019-head}, machine translation \cite{ng2019facebook}, question answering \cite{yang2019xlnet} \textit{inter alia}.
Transformers also form the backbone of state-of-the-art pre-trained language models, e.g. BERT \cite{Devlin:2018}, GPT-2 \cite{radford2019language} and GPT-3 \cite{Larochelle2020}, 
that have further boosted performance on various data-driven NLP problems.

The key ingredient in the Transformer architecture is the multi-head attention mechanism, which is an assembly of multiple attention functions \cite{bahdanau2015neural} applied in parallel.
In practice, each attention head works independently, which allows the heads to capture different kinds of linguistic phenomena \cite{clark2019does,goldberg2019assessing,ettinger2019bert,jawahar2019does}.
A natural question arises in this context: How many heads does a transformer need?\looseness=-1

\newcite{Michel:2019} offers the insight that \textit{a large portion of the Transformer's heads can be pruned without significantly degrading the test accuracy on the desired task}.
The experimental evidence behind their claim is a simple greedy procedure that sequentially removes heads.
This suggests that a better pruner could reveal that a much larger portion of the heads can be safely removed.
To provide a more robust answer to \citeauthor{Michel:2019}'s question, we build a high-performance pruner and show that their approach itself significantly underestimates the number of Transformer heads that can be pruned away.\looseness=-1

\begin{figure*}
\centering
  \begin{subfigure}[b]{.5\textwidth}
    \centering
    \includegraphics[width=\textwidth]{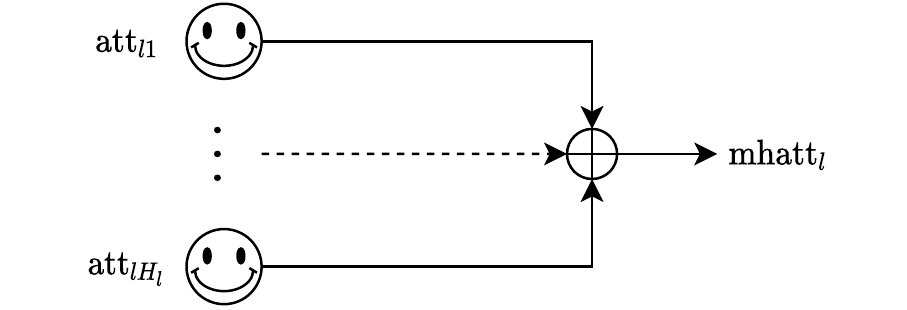}
    \caption{Standard multi-head attention. Attention heads in the same layer $l$ are combined by summation. }
    \label{fig:standard}
  \end{subfigure}
  \begin{subfigure}[b]{\textwidth}
    \centering\includegraphics[width=\textwidth]{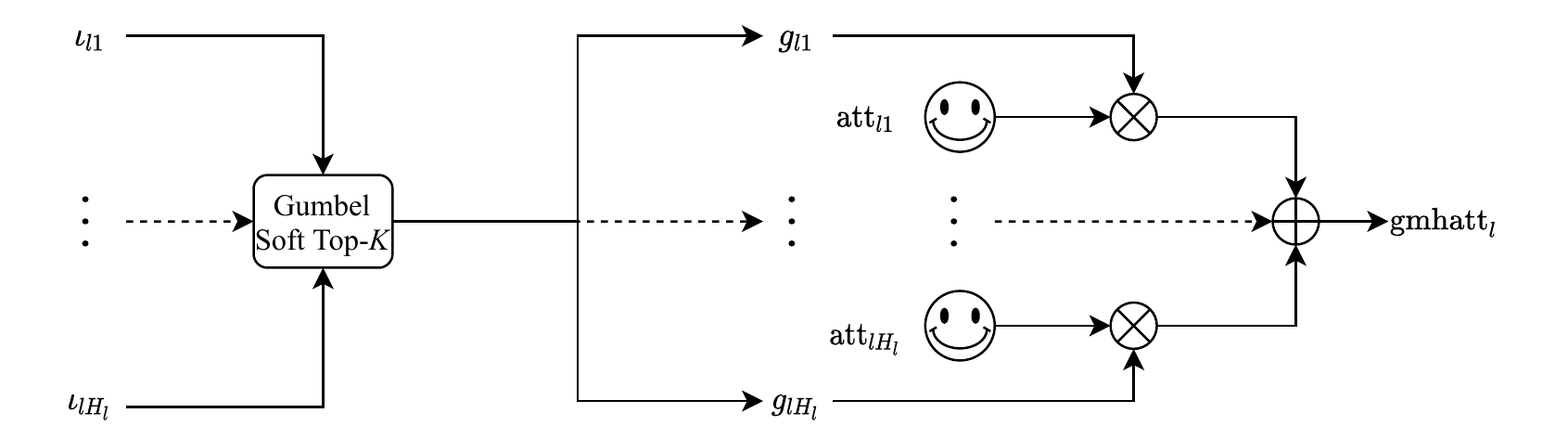}
    \caption{Multi-head attention with differentiable subset pruning. A gate $g_{lh}\in [0,1]$ is applied on each head. The gate values are determined by Gumbel Soft top-$K$ algorithm, which takes the head importance scores $\iota_{lh}$ as inputs. The least "important" heads will have their gate values close to $0$, so they are pruned in effect. }
    \label{fig:gumbel}
  \end{subfigure}
  \caption{Illustration of gated multi-head attention compared with standard multi-head attention.}
  \label{fig:diagram}
\end{figure*}

From a bird's eye view, our paper contributes the proposal that Transformer head pruning is best viewed as a \defn{subset selection} problem.
Subset selection is common across many areas of NLP: from extractive summarization \cite{gillenwater-etal-2012-discovering} to vowel typology \cite{cotterell-eisner-2017-probabilistic}.
In the case of head pruning, the concrete idea is that the user specifies a number of heads $K$ that they would like their Transformer to have depending on their budgetary and other constraints, and then the pruner enforces this constraint.
Methodologically, we present a differentiable subset pruner (\cref{fig:diagram}) that makes use of Gumbel machinery; specifically the Gumbel top-$K$ procedure of \newcite{vieira2014gumbel}.
This construction allows us to relax our pruner into a differentiable sampling routine 
that qualitatively resembles a discrete analogue of dropout \cite{Srivastava:2014,pmlr-v48-gal16}.

Empirically, we perform experiments on two common NLP tasks: natural language inference \cite[MNLI;][]{Williams:2018} and machine translation \cite[IWSLT2014;][]{cettolo2014report}. 
We show that our differentiable subset pruning scheme outperforms two recently proposed Transformer head pruners---\newcite{Michel:2019} and \newcite{Voita:2019}---on both tasks
in terms of sparsity--performance trade-off. 
Our method recovers a pruned Transformer that has $\approx 80\%$ accuracy on MNLI and $\approx 30$ BLEU score on IWSLT when more than $90\%$ of the heads are removed, which brings about $\approx 33\%$ inference speedup and $\approx 24\%$ model size shrinkage.\footnote{See \cref{sec:efficiency}.} 

Our experiments also suggest several broader conclusions about pruning Transformers. 
In this paper, we taxonomize existing pruning methods into two pruning paradigms: pipelined pruning and joint pruning. 
\defn{Pipelined pruning} consists of two stages: i) training or fine-tuning an over-parameterized model on the target task and ii) pruning the model after training. 
A number of techniques fall into this category \cite{lecun, Hassibi:93, han2015deep,Molchanov:2019}.
In contrast, \defn{joint pruning} blends the pruning objective into the training objective by training or fine-tuning the over-parameterized model with a sparsity-enforcing regularizer, sometimes followed up by a trivial post-processing step to arrive at a final sparse model. \newcite{Kingma:2015} and  \newcite{louizos2017learning}  are examples of this kind of pruning. 
We show that pipelined head pruning schemes, such as that of \citeauthor{Michel:2019}, underperform compared to joint head pruning schemes, such as that of \newcite{Voita:2019}.
Our differentiable subset pruner can be adapted to both paradigms and it outperforms prior work in both, especially in high sparsity regions.
\looseness=-1

\section{Background: Multi-head Attention}
In this section, we provide a detailed overview of multi-head attention \cite{vaswani2017attention} in order to develop the specific technical vocabulary to discuss our approaches for head pruning. 
We omit details about other parts of the Transformer and refer the reader back to the original work of \citet{vaswani2017attention}. 
First, let $\zz = z_1, \ldots, z_T$ be a sequence of $T$ real vectors where $z_t \in \R^{d}$, and let $q \in \R^{d}$ be a query vector.
An \defn{attention mechanism} is defined as
\begin{equation}
    \att(\zz, q) = W_o \sum_{t=1}^T \alpha_t(q) W_v z_t
\end{equation}
\quad where
\begin{equation}
    \alpha_t(q) = \softmax\left(    \frac{q^{\top} W_q^{\top} W_k z_t}{\sqrt{d}}\right)_t
\end{equation}
The projection matrices $W_o, W_v, W_q, W_k \in \R^{d \times d}$ 
are learnable parameters.
In self-attention, query $q$ comes from the same sequence $\zz$.

A Transformer is composed of $L$ identical layers. 
In layer $1 \leq l \leq L$, $H_l$ different attention mechanisms are applied in parallel; importantly, it is this parallelism that has lead to the rise of the Transformer---it is a more efficient architecture in practice so it can be trained on more data.
Each individual attention mechanism is referred to as a \defn{head}; thus, \defn{multi-head attention}
is the simultaneous application of multiple attention heads in a
single architecture. 
In \newcite{vaswani2017attention}, the multiple heads are combined through summation:
\begin{equation}
 \mhatt_l(\zz, q) =  \sum_{h=1}^{H_l} \att_{lh}(\zz, q)
\end{equation}
where $\att_{lh}$ is the $h^{\text{th}}$ attention head in the $l^{\text{th}}$ layer.
We also introduce a \defn{gate variable} $g_{lh}$ that takes values in the interval $[0, 1]$:
\begin{equation}\label{eq:gmhatt}
 \gmhatt_l(\zz, q) = \sum_{h=1}^{H_l} g_{lh} \cdot \att_{lh}(\zz, q)
\end{equation}
Inserting $g_{lh}$ into the multi-head attention enables our pruning approach: setting the gate variable to $g_{lh}=0$ means the head $\att_{lh}$ is pruned away. 

In the following sections, for the sake of notational simplicity, we ignore the layer structure of heads and label heads with a single index $h\in\{1,\ldots,H\}$, where $H = \sum_{l=1}^L H_l$ is the total number of heads in the unpruned model. 

\section{Differentiable Subset Pruning}
\label{sec:Gumbel}%
In this section, we propose a new head pruning technique that we term \defn{differentiable subset pruning}.
The key insight behind our approach is that head pruning can be viewed as subset selection. 
Concretely, our goal is to find a subset of $K$ heads (where $K$ is a user-specified positive integer) that still allows the model to achieve high performance. 
Many neural network pruners, e.g. \newcite{Voita:2019}'s proposed head pruning technique, make it notably difficult to pre-specify the number of pruned heads $K$\footnote{Later discussed in \cref{sec:exp_joint}}.
To make our subset pruner differentiable,
we apply the Gumbel--$\softmax$ trick \cite{maddison2017concrete} and its extension to subset selection \cite{vieira2014gumbel,xie2019reparameterizable}. 
This gives us a pruning scheme that always returns the specified number of heads and can be applied in a pipelined or a joint setting.
In both cases, the differentiability is necessary to learn the head weights.\looseness=-1

\subsection{Background: Gumbel-(soft)max}\label{sec:gumbel-background}
Let $\calH=\{1, \ldots, H\}$ be the set of Transformer heads in a given architecture.
Our goal is to return a subset of heads $\calJ \subseteq \calH$ where $|\calJ| = K$
for any user-specified value of $K$.
We use the notation $\iota_h > 0$ to denote a head \defn{importance score} of the specific head $h$. 
The head importance score intuitively corresponds to how much we would like to have the head $h$ in the subset of heads $\calJ$.

We start our exposition by reviewing the Gumbel trick in the context of selecting a single head ($K=1$) and then move onto discussing its extension to subset selection.
Given the head importance scores $\iota_h$, suppose we would like to sample a subset $\calJ$ of size $1$ according to the following distribution
\begin{equation}
    p(\calJ = \{h\}) = \frac{\iota_h}{Z} \propto \iota_h
\end{equation}
where $Z=\sum_{h=1}^H \iota_h$ is the normalization constant. 
The simplest way to achieve this to use standard  categorical sampling.
However, as has been noted by \newcite{maddison2014sampling}, categorical sampling is not differentiable.
Luckily, there is a two-step process to massage categorical sampling into a differentiable sampling procedure: 1) reparameterize the categorical using Gumbels and 2) soften the $\argmax$ into a $\softmax$. 

\subsubsection{Step 1: Reparameterization}\label{sec:gumbel-reparam}
We can reparameterize categorical sampling using the Gumbel-max trick \cite{Gumbel:1954} to first separate the sampling from the parameter
that we wish to differentiate with respect to.
The idea of the Gumbel max trick is that categorical sampling can be viewed as a perturb-and-max method.
If we first perturb the logits $\log (\iota_h)$ with Gumbel noise $n_h\sim \text{Gumbel}(0,1)$ such that $r_h=\log(\iota_h)+n_h$, 
then sampling from a categorical is equivalent to taking an $\argmax$:
\begin{equation}
    h^* = \argmax_{h\in\calH} r_h
\end{equation}
Were $\argmax$ differentiable, we would be done; unfortunately it is not.

\subsubsection{Step 2: Relaxing the $\argmax$}\label{sec:relax-gumbel}
Now to construct a fully differentiable procedure, we replace the $\argmax$ with a $\softmax$.
The intuition here is that the output of $\argmax$ may be viewed as an one-hot vector
with the one corresponding to the index
of the $\argmax$.\footnote{More precisely, $\argmax$ returns a set.
In our terminology, it would return a multi-hot vector.
We ignore this case in our exposition for simplicity.}
The insight, then, is to relax the one-hot vector output by the $\argmax$ into a $\softmax$ as follows:
\begin{equation}\label{eq:gumbel-softmax}
    g_h = \frac{\exp(r_h)}{\sum_{h'=1}^H \exp(r_{h'})}
\end{equation}
This technique is called the Gumbel-\emph{$\softmax$} trick \cite{jang2016categorical},
and the resulting distribution is known as the Concrete distribution \cite{maddison2017concrete}.\footnote{Using the Gumbel-$\softmax$ results in a biased estimate of the gradient.
Subsequent work removed this bias \cite{NIPS2017_ebd6d2f5}.}
It is often desirable to add an additional annealing parameter  $\tau > 0$ to the Gumbel-$\softmax$:
\begin{equation}
    g_h = \frac{\exp \left(r_h/\tau \right)}{\sum_{h'=1}^H \exp \left(r_{h'}/\tau \right)}
\end{equation}
As the temperature tends to zero, i.e. $\tau \rightarrow 0$, the $\softmax$ turns into the $\argmax$.
Thus, through the tunable $\tau$, we can arbitrarily approximate the $\argmax$ as a differentiable function.
    
\subsection{Differentiable Subset Selection}
The Gumbel trick can be generalized to cases where we wish to sample an entire set of heads.
This is called the Gumbel-top-$K$ trick.
The idea is that, rather than simply taking the max, we sort and the take the top-$K$ largest perturbed logits \cite{YELLOTT1977109,vieira2014gumbel, kool2019stochastic}.
One way to think of the algorithm is that we are repeating the Gumbel trick $K$ times until we have the desired number of heads.
Following the exposition in \cref{sec:gumbel-background}, we 
divide our discussion into two sections.

\subsubsection{Step 1: Reparameterization}\label{sec:gumbel-topk}
Similar to the top-$1$ case, we
start by sampling the first head using
the perturb-and-max strategy:
\begin{equation}
\label{eq:hard_k}
    h_1^* = \argmax_{h\in\calH} r_h
\end{equation}
Then we remove $h_1^*$ from the pool of heads under consideration and repeat the same procedure:
\begin{align}
    h_2^* &= \argmax_{h \in \calH \setminus \{h_1^*\}} r_h \\
   & \quad\quad\quad  \vdots  \nonumber \\ 
    h_K^* &= \!\!\!\!\!\!\!\!\!       \argmax_{h \in \calH \setminus \{h_1^*, \ldots, h_{K-1}^*\}} r_h
    \label{eq:hard_k_permute}
\end{align}
The probability of sampling these heads \emph{in this order} is given by the following expression:
\begin{align}
    p(h_1^*, \ldots, h_K^*) = \frac{\iota_{h^*_1}}{Z} \cdots \frac{\iota_{h^*_K}}{Z-\sum_{k=1}^{K-1}\iota_{h_k^*}}
\end{align}
Thus, the probability of a set $\calJ$ is given by
\begin{align}\label{eq:perm}
    p(\calJ = \{h_1^*, \ldots, &h_K^*\})  \\
    &=\sum_{\pi \in \mathbb{S}_k} p(h_{\pi_1}^*, \ldots, h_{\pi_K}^*) \nonumber
\end{align}
where $\mathbb{S}_K$ is the set of all permutations of $K$ items. 
This is hard to compute as it involves a sum over permutations. 
For a detailed discussion on computing \cref{eq:perm}, we refer the reader to the discussion in \newcite{vieira2021order} and \newcite{vieira2021smallest}. 
Ultimately, however, computing the exact probability of a subset of heads $\calJ$ is unnecessary for this approach.

As an aside, we note that this procedure is equivalent to a differentiable version of the classical reservoir sampling algorithm \cite{reservoir}. 

\subsubsection{Step 2: Relaxing the $\argmax$}
The Gumbel-top-$K$ trick can be relaxed similarly to the top-1 case. This was first shown in detail by \newcite{xie2019reparameterizable}. 
Here, we provide a detailed overview of the algorithm by analogy to the top-1 case.
Similarly, the output of Gumbel-top-$K$%
can be viewed as a $K$-hot vector, which is the sum of the $K$ one-hot vectors produced in \cref{eq:hard_k}--\cref{eq:hard_k_permute}.
As before, we begin by relaxing the one-hot vector
of the first head:\looseness=-1
\begin{equation}\label{eq:soft_k}
    g^{(1)}_h = \frac{\exp(r^{(1)}_h/\tau)}{\sum_{h'=1}^H \exp(r^{(1)}_{h'}/\tau)}
\end{equation}
This is a straight-forward analogue of the $\argmax$ relaxation discussion in \cref{sec:relax-gumbel}.
Next, we continue relaxing the successive $\argmax$es with successive $\softmax$es \cite{plotz2018neural} as follows:
\begin{align}
    g_h^{(2)} &= \frac{\exp(r^{(2)}_h/\tau)}{\sum_{h'=1}^H \exp(r^{(2)}_{h'}/\tau)} \\
   & \quad\quad\quad\quad\quad \vdots  \nonumber \\ 
    g^{(K)}_h &= \frac{\exp(r^{(K)}_h/\tau)}{\sum_{h'=1}^H \exp(r^{(K)}_{h'}/\tau)}
    \label{eq:soft_k_permute}
\end{align}
where the $r_h^{(k)}$ are defined recursively 
\begin{align}\label{eq:recursion}
    r_h^{(1)} &= r_h \\
    r_h^{(k+1)} &= r_h^{(k)} + \log\left(1 - g_h^{(k)}\right) &
\end{align}

\newcite{xie2019reparameterizable} argue that the above recursion corresponds to a reasonable relaxation of the Gumbel-top-$K$ trick presented in \cref{sec:gumbel-topk}.
To understand the motivation behind the recursion in \cref{eq:recursion},
note that if $g_h^{(k)}=1$, which would happen if the head has been sampled, i.e. no relaxation, then that head would not be selected again as we have $r_h^{(k+1)} = -\infty$.
As the scheme is a relaxation of hard sampling, we will not have $g_h^{(k)}=1$ as long as $r_h^{(k)}$ is finite and $\tau > 0$. 
Thus, the procedure corresponds to something akin to a soft sampling.

Finally, we sum over all the relaxed one-hot vectors $g_h^{(k)}$ in \cref{eq:soft_k}--\cref{eq:soft_k_permute}
to arrive at our softened $K$-hot gate:\looseness=-1
\begin{equation}\label{eq:final-gate}
    g_h = \sum_{k=1}^K g_h^{(k)}
\end{equation}
It is \cref{eq:final-gate} that we finally plug into the gated attention mechanism presented in  \cref{eq:gmhatt}.

\subsection{Training the Subset Pruner}
The differentiable subset pruning approach can be applied in either a pipelined or a joint pruning setting. 
(Please refer back to the last paragraph of \cref{sec:introduction} for a discussion of the two different settings.)
Our approach is parameterized identically in both settings, however.
Specifically, we define head importance score as follows:
\begin{equation}
    \iota_h = \exp(w_h)
\end{equation}
where $w_h$ is the $h^{\text{th}}$ component of a vector of real-valued head weights $\mathbf{w} \in \mathbb{R}^H$. 
In our setting, the distinction between pipelined pruning and joint pruning is relatively trivial.
In the pipelined setting, we learn the head importance weights $\mathbf{w}$ for a model that has been trained on the task and leave the model parameters untouched.
On the other hand, in the joint setting,
we simultaneously learn the head importance weights and the model parameters.
In this regard, our differentiable subset pruner much more closely resembles \newcite{Voita:2019}'s method in that we \emph{learn} head-specific importance weights. 
On the other hand, \newcite{Michel:2019}'s method makes use of an unlearned gradient-based importance measure. 
In contrast to \citeauthor{Voita:2019}, however, our differentiable subset pruner ensures that it returns a specific pre-specified number of heads.\looseness=-1

\section{Experiments}
\subsection{Model and Data}
We investigate two Transformer-based models in the empirical portion of the paper. 
\paragraph{$\bert$.} BERT \cite[Bidirectional Encoder Representations from Transformers;][]{Devlin:2018} is essentially a Transformer encoder. Since there is no decoder part, BERT only has self-attention. We focus on the \texttt{base-uncased} model with 12 layers and 12 heads in each layer (144 heads in total). We use the implementation of Hugging Face \cite{wolf-etal-2020-transformers}. The model is pre-trained on large text corpora using masked language modeling (MLM) and next sentence prediction (NSP).
We fine-tune BERT on the Multi-Genre Natural Language Inference \cite[MNLI;][]{Williams:2018} corpus. 
The hyper-parameters are tuned on the "matched" validation set, and accuracy is reported on the "mismatched" validation set.%

\paragraph{$\ed$.} We implement a Transformer-based encoder--decoder  model with 6 encoder layers, 6 decoder layers and 6 heads in each layer (72 heads in total). The model has three types of attention heads: encoder self-attention, decoder self-attention, and encoder--decoder cross attention. 
We use the \texttt{fairseq} toolkit \cite{ott-etal-2019-fairseq} for our implementation.
We train the model on the International Workshop on Spoken Language Translation \cite[IWSLT2014;][]{cettolo2014report} German-to-English dataset. The hyper-parameters are tuned on the validation set, and 4-gram BLEU scores computed with \texttt{multi-bleu.perl} \cite{koehn-etal-2007-moses} are reported on the held-out test set. We use beam search with a beam size set to 5 for decoding.\looseness=-1 

\subsection{Baselines}
We compare our approach to pruners in both the pipelined and the joint paradigms.  
We refer to the pipelined version of our differentiable subset pruning as \textbf{pipelined DSP}
and to the joint version as \textbf{joint DSP}.
Our specific points of comparison are listed below.
\subsubsection{\citeauthor{Michel:2019}}
\citeauthor{Michel:2019} follows the pipelined pruning paradigm. Concretely, given a dataset $\calD = \{(\yy_m, \xx_m)\}_{m=1}^M$, the importance of a head is estimated with a gradient-based proxy score \cite{Molchanov:2019}:
\begin{equation}\label{eq:importance}
\iota_{h} = \frac{1}{M} \sum_{m=1}^M \left| \frac{\partial \calL(\yy_m, \xx_m) }{\partial g_{h}} \right| \geq 0,
\end{equation}
where $\calL$ is the task-specific loss function. 
Then, all the heads in the model are sorted accordingly and removed one by one in a greedy fashion.
The importance scores are re-computed every time a certain number of heads are removed. %

\subsubsection{\citeauthor{Voita:2019}} 
In the fashion of joint pruning, \citeauthor{Voita:2019} applies a stochastic approximation to $L_0$ regularization \cite{louizos2017learning} to the gates to encourage the model to prune less important heads. 
The gate variables are sampled from a binary Hard Concrete distribution \cite{louizos2017learning} independently, parameterized by $\phi_h$. The $L_0$ norm was relaxed into the sum of probability mass of gates being non-zero:
\begin{equation}
    L_C(\phi)=\sum_{h=1}^H (1-P(g_h=0|\phi_h)),
\end{equation}
which was then added to the task-specific loss $\calL$:
\begin{equation}
    \calR(\theta, \phi) = \calL(\theta, \phi) + \lambda L_C(\phi),
\end{equation}
where $\theta$ are the parameters of the original model, and $\lambda$ is the weighting coefficient for the regularization, which we can use to indirectly control the number of heads to be kept. 

\subsubsection{Straight-Through Estimator (STE)}
In this baseline, the Gumbel soft top-$K$ in joint DSP is replaced with hard top-$K$, while the hard top-$K$ function is back-propagated through as if it had been the identity function, which is also termed as straight-through estimator \cite{DBLP:journals/corr/BengioLC13}.

\subsubsection{Unpruned Model} 
The model is trained or fine-tuned without any sparsity-enforcing regularizer and no post-hoc pruning procedure is performed.
We take this comparison to be an upper bound on the performance of any pruning technique.

\subsection{Experimental Setup}
\paragraph{Pipelined Pruning.}
For the two pipelined pruning schemes, the model is trained or fine-tuned on the target task (3 epochs for $\bert$ and 60 epochs for $\ed$) %
before being pruned. 
We learn the head importance weights for pipelined DSP for one additional epoch in order to have
an apples-to-apples comparison with \citeauthor{Michel:2019} in terms of compute (number of gradients computed).%
\paragraph{Joint Pruning.}
The model is trained or fine-tuned for the same number of epochs as pipelined pruning while sparsity-enforcing regularization is applied.
We found it hard to tune the weighting coefficient $\lambda$ for \citeauthor{Voita:2019} to reach the desired sparsity (see \cref{sec:exp_joint} and \cref{fig:step}). 
For the ease of comparison with other approaches, we adjust the number of unpruned heads to the targeted number by re-including heads with the highest gate values from the discarded ones, or excluding those with the smallest gate values in the kept ones. We make sure the adjustments are as small as possible.

\paragraph{Annealing Schedule.}
In our experiments, we choose a simple annealing schedule for DSP where the temperature $\tau$ cools down in a log-linear scale within a predefined number of steps $N_{\text{cooldown}}$ from an initial temperature $\tau_{\text{ini}}$ and then stays at the final temperature $\tau_{\text{end}}$ for the rest of the training steps:
\begin{align}
    &\log \tau 
    = \log \tau_{\text{ini}}\,\,- \\ 
    &\quad\quad \min\left\{\frac{n}{N_{\text{cooldown}}},1\right\} \cdot \Big(\log \tau_{\text{ini}} - \log \tau_{\text{end}}\Big) \nonumber
\end{align}
where $n$ is the number of training steps that has been run. We report the set of hyperparameters used in our experiments in \cref{app:hyper}.

\subsection{Results}
The test performance under various sparsity levels obtained by multiple pruning methods are presented in \cref{fig:mnli}, \cref{fig:mt} and \cref{app:results}. 
We also zoom in to results when more than two-third of the heads are pruned in \cref{fig:mnli_zoom} and \cref{fig:mt_zoom}, where the differences between the various methods are most evident. 

\addtocounter{figure}{1}
\begin{figure}
\centering
  \includegraphics[width=\columnwidth]{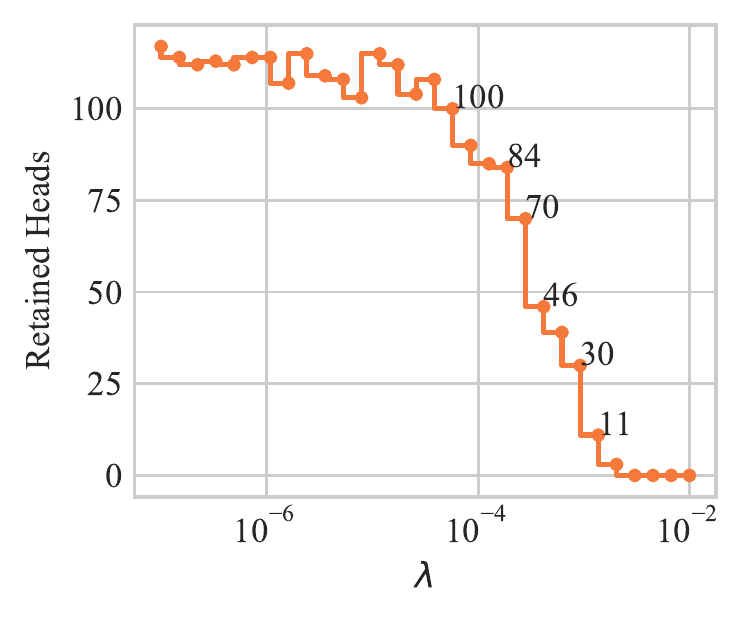}
  \caption{Number of unpruned heads as a function of $L_0$ regularization coefficient $\lambda$ for \citeauthor{Voita:2019}.}
  \label{fig:step}
\end{figure}

\begin{figure}[t]
\centering
  \includegraphics[width=\columnwidth]{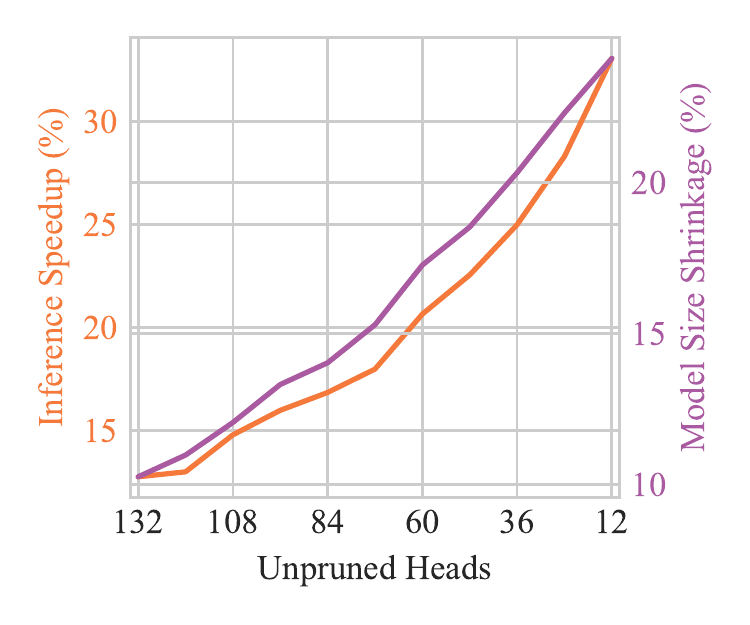}
  \caption{Inference speedup ($\%$) and model size shrinkage ($\%$) of pruned $\bert$ model on the MNLI-mismatched validation set as a function of remaining heads.\looseness=-1}
  \label{fig:compression}
\end{figure}

\addtocounter{figure}{-3}
\begin{figure*}
\centering
  \begin{subfigure}[b]{.49\textwidth}
    \centering\includegraphics[width=\columnwidth]{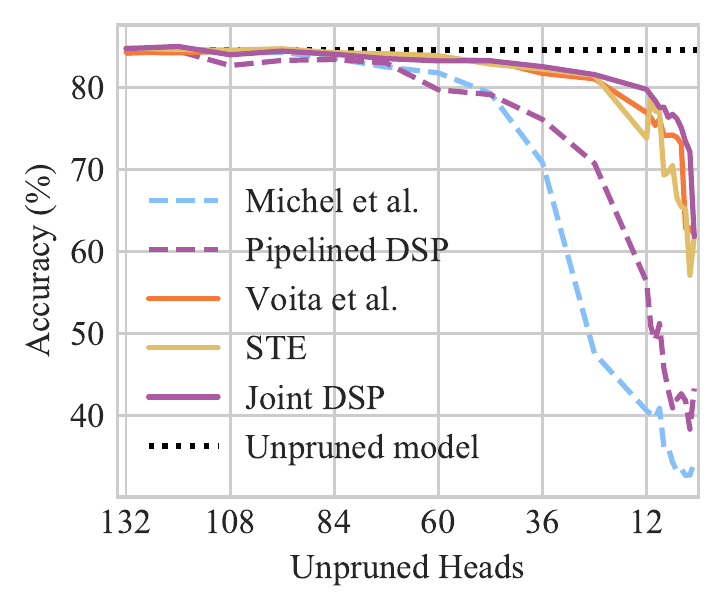}
    \caption{Accuracy on the MNLI-mismatched validation set as a function of number of unpruned heads in $\bert$.}
    \label{fig:mnli}
  \end{subfigure}
  \begin{subfigure}[b]{.49\textwidth}
    \centering\includegraphics[width=\columnwidth]{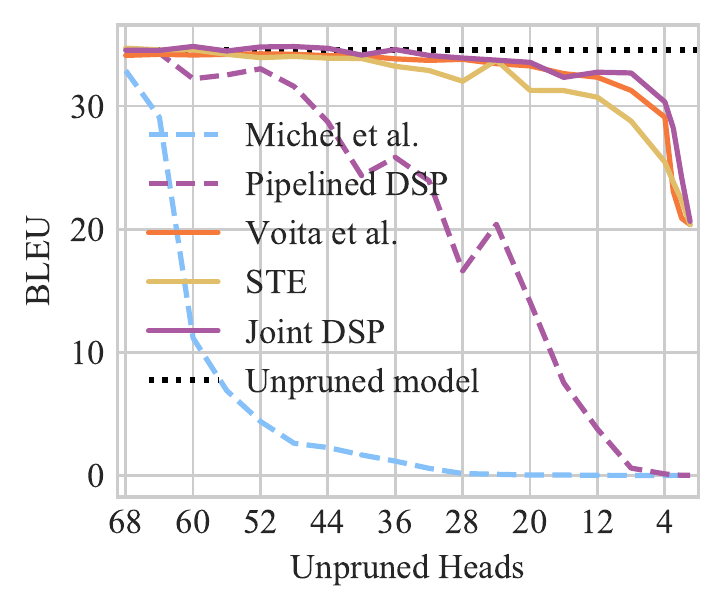}
    \caption{BLEU score on IWSLT test set as a function of number of unpruned heads in $\ed$.}
    \label{fig:mt}
  \end{subfigure}
    \newline
    \vspace{0cm}
    \newline
  \begin{subfigure}[b]{.49\textwidth}
    \centering\includegraphics[width=\columnwidth]{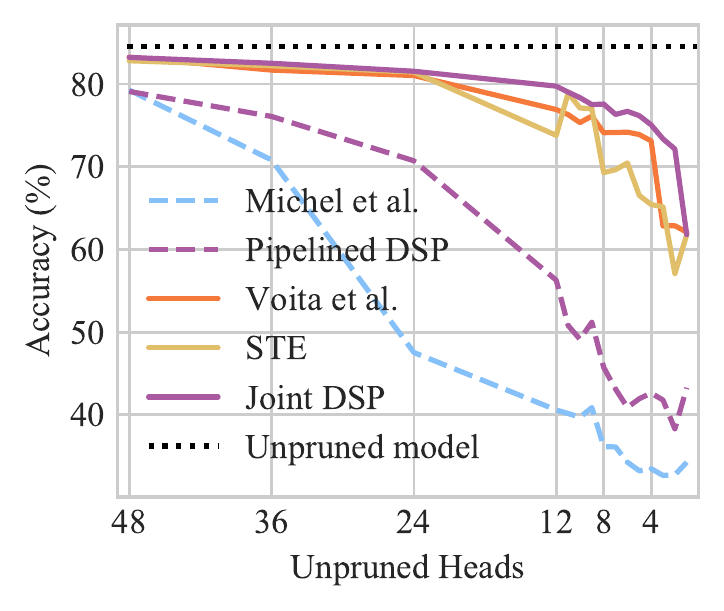}
    \caption{We zoom in on the portion of \cref{fig:mnli} where few heads remain unrpuned.}
    \label{fig:mnli_zoom}
  \end{subfigure}
  \begin{subfigure}[b]{.49\textwidth}
    \centering\includegraphics[width=\columnwidth]{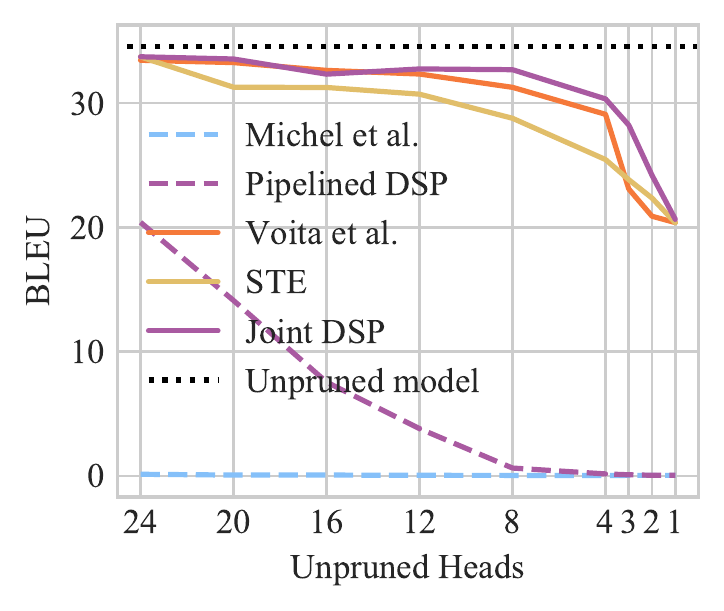}
    \caption{We zoom in on the portion of \cref{fig:mt} where few heads remain unrpuned.}
    \label{fig:mt_zoom}
  \end{subfigure}
  \caption{A comparison of various pruning methods.}
  \label{fig:comparison}
\end{figure*}

\addtocounter{figure}{2}
\begin{figure*}
\centering
  \begin{subfigure}[b]{.49\textwidth}
    \centering\includegraphics[width=\textwidth]{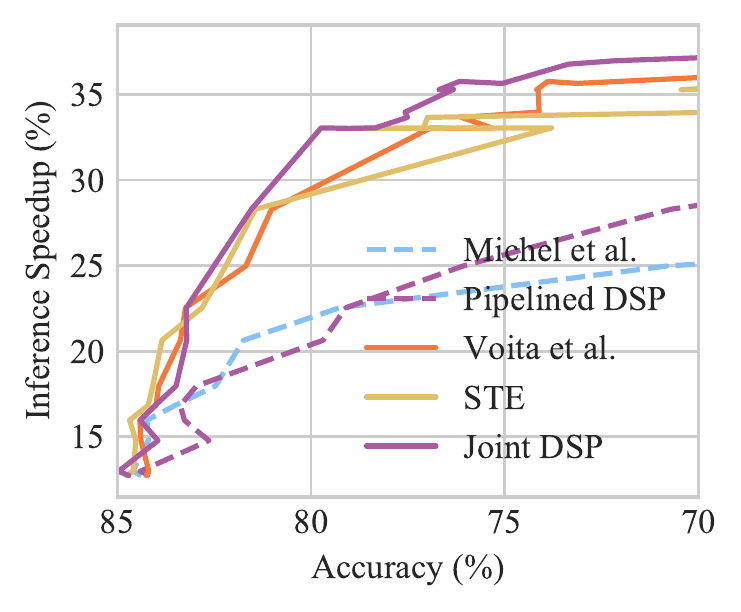}
    \caption{Inference speedup as a function of accuracy.}
    \label{fig:speedup}
  \end{subfigure}
  \begin{subfigure}[b]{.49\textwidth}
    \centering\includegraphics[width=\textwidth]{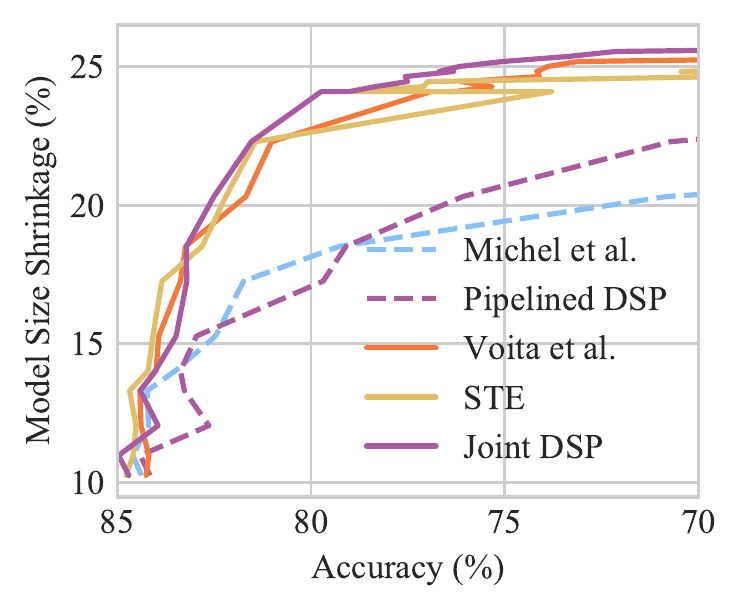}
    \caption{Model size shrinkage as a function of accuracy.}
    \label{fig:shrinkage}
  \end{subfigure}
  \caption{Inference speedup ($\%$) and model size shrinkage ($\%$) of the various pruned $\bert$ models vs. accuracy ($\%$) on the MNLI-mismatched validation set. }
  \label{fig:compression_detail}
\end{figure*}

\section{Discussion}
\subsection{Pipelined Pruning}
We first compare the two pipelined pruning methods: \newcite{Michel:2019} and pipelined DSP. 
As shown in \cref{fig:comparison}, pipelined DSP outperforms \citeauthor{Michel:2019} by a large margin. 
For example, on the MNLI task, when there are $24$ heads left in the model, pipelined DSP keeps an accuracy above $70\%$, but \citeauthor{Michel:2019} drops below $50\%$.
On the IWSLT dataset, when only $24$ heads are left unpruned, the $\ed$ 
pruned with \citeauthor{Michel:2019} cannot produce meaningful outputs ($\approx 0$ BLEU score), while pipelined DSP achieves higher than $20$ BLEU. 
The results indicate that the learned head importance scores are more useful for pruning than those computed with gradient-based measures. 

\begin{figure*}
\centering
  \begin{subfigure}[b]{.49\textwidth}
    \centering\includegraphics[width=\textwidth]{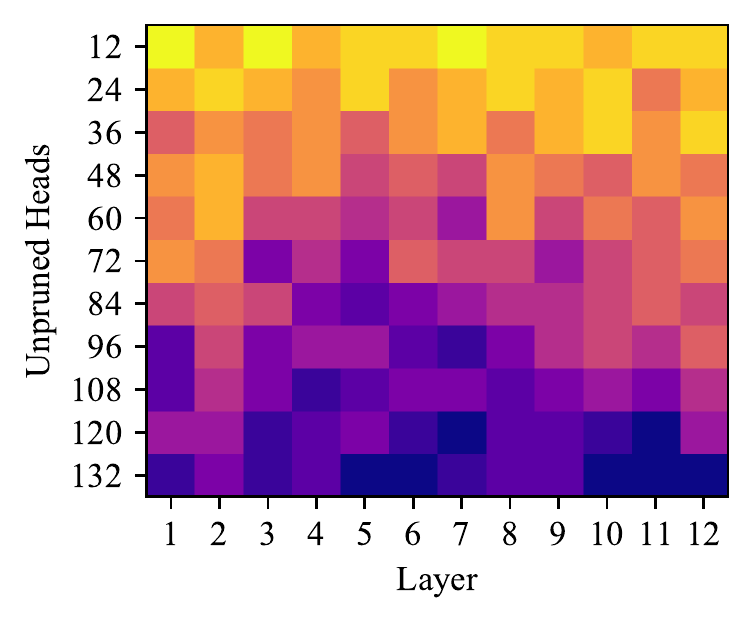}
    \caption{$\bert$}
    \label{fig:distribution_mnli}
  \end{subfigure}
  \begin{subfigure}[b]{.49\textwidth}
    \centering\includegraphics[width=\textwidth]{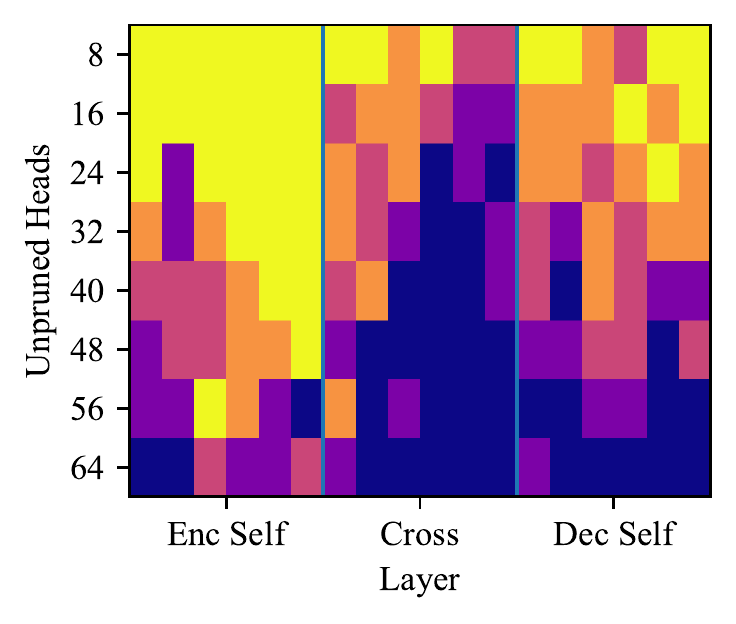}
    \caption{$\ed$}
    \label{fig:distribution_mt}
  \end{subfigure}
  \caption{Distribution of unpruned heads across layers. Darkness of the color increases monotonically with the number of heads.\looseness=-1}
  \label{fig:distribution}
 \vspace{10pt}
\end{figure*}

\subsection{Joint Pruning}
\label{sec:exp_joint}
We then compare the three joint pruning methods: \newcite{Voita:2019}, STE and joint DSP. 
Impressively, joint DSP is able to prune up to $91.6\%$ (12 heads left) and $94.4\%$ (4 heads left) of heads in $\bert$ and the $\ed$ respectively without causing much degradation in test performance ($5.5\%$ drop in accuracy for MNLI and $4.22$ drop in BLEU score for IWSLT).
\citeauthor{Voita:2019} and STE are neck and neck with joint DSP when the model is lightly pruned, but joint DSP gains the upper hand when less than $\sfrac{1}{6}$  of the heads are left unpruned. 

In addition, with \citeauthor{Voita:2019}'s method, it is much harder to enforce a hard constraint on the number of unpruned heads. 
This difficulty is intrinsic to their method as \citeauthor{Voita:2019}'s method relies on the regularization coefficient $\lambda$ to indirectly control the sparsity. 
In practice, our experiments indicate that $\lambda$ is hard to tune and there are certain levels of sparsity that cannot be reached. 
The difficulty in tuning $\lambda$ is shown in \cref{fig:step}; we see that the number of unpruned heads does not decrease monotonically as $\lambda$ increases; on the contrary, it often fluctuates. 
There also appears to be an upper bound ($117$) on the number of heads that can be kept no matter how small $\lambda$ is. 
More importantly, a small increase in $\lambda$ can sometimes drastically reduce the number of heads. 
For instance, when $\lambda$ is increased from $0.0009$ to $0.0014$, the number of heads reduced quickly from $30$ to $11$.
Therefore, we conclude that \citeauthor{Voita:2019}'s method is inadequate if the user requires a pre-specified number of Transformer heads. %
In contrast, DSP (as well as STE), our proposal,  enables us to directly specify the number of heads we want to keep in accordance with our computation budget.

\subsection{Pipelined Pruning vs Joint Pruning}
\label{sec:comparison}
Lastly, we offer a philosophical comparison of the two pruning paradigms. 
It is clear from \cref{fig:comparison} that the joint pruning methods are superior to pipelined pruning methods for both tasks, as models sparsified with the joint pruning schemes (joint DSP, STE and \citeauthor{Voita:2019}) perform better than those pruned with pipelined schemes (pipelined DSP and \citeauthor{Michel:2019}) under almost every sparsity level. %
This suggests that joint training is more effective in finding sparse subnetworks than pipelined pruning. %
Moreover, joint pruning is also more computationally efficient. In addition to the same number of epochs required by both paradigms for  training/fine-tuning, pipelined pruning requires us to learn or estimate gradient-based head importance scores for one extra epoch. 
Even though joint pruning methods train $H$ more parameters during training/fine-tuning, $H$ is typically orders of magnitudes smaller than the total number of model parameters, so the additional computational overhead is negligible.

\subsection{Inference Efficiency}
\label{sec:efficiency}
In this section, we obtain the pruned model by actually removing the heads with mask values $0$. Empirically, we observe substantial wallclock improvements in our pruned models compared to unpruned models. 
In practice, we found that the inference efficiency improves monotonically as the number of unpruned heads decrease and is not significantly impacted by the distribution of heads across layers. Taking $\bert$ on MNLI-mismatched validation set (batch size of 8) as an example, we randomly sample $10$ head masks for each sparsity level, measure their inference speedup and model size shrinkage compared to the unpruned model, and report the average in \cref{fig:compression}. In general, head pruning does lead to a faster and smaller model, and the more we prune, the faster and smaller the model becomes.\looseness=-1

Comparison of various pruning schemes is displayed in \cref{fig:compression_detail}. If we set a threshold for accuracy (e.g. 80\%), joint DSP returns a model with a $\approx 33\%$ speedup in execution time and $\approx 24\%$ decrease in model size.\looseness=-1

\begin{figure*}
\centering
  \begin{subfigure}[b]{.49\textwidth}
    \centering\includegraphics[width=\textwidth]{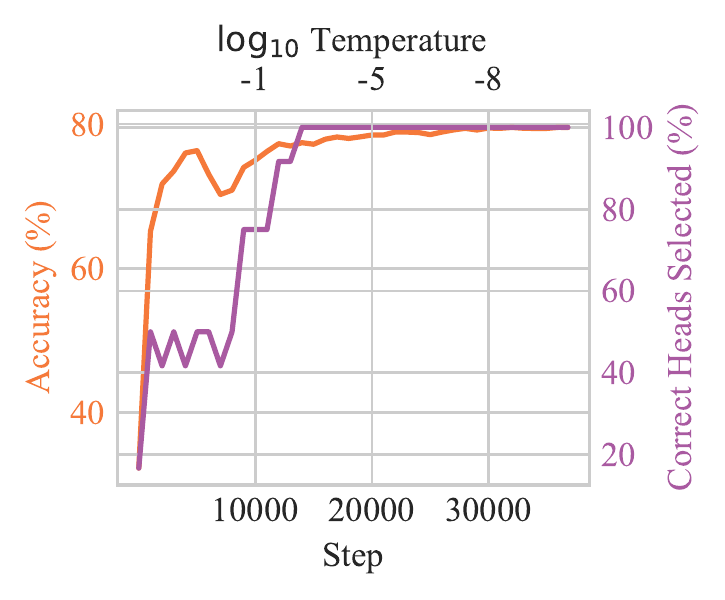}
    \caption{With Annealing}
    \label{fig:dynamic_annealing}
  \end{subfigure}
  \begin{subfigure}[b]{.49\textwidth}
    \centering\includegraphics[width=\textwidth]{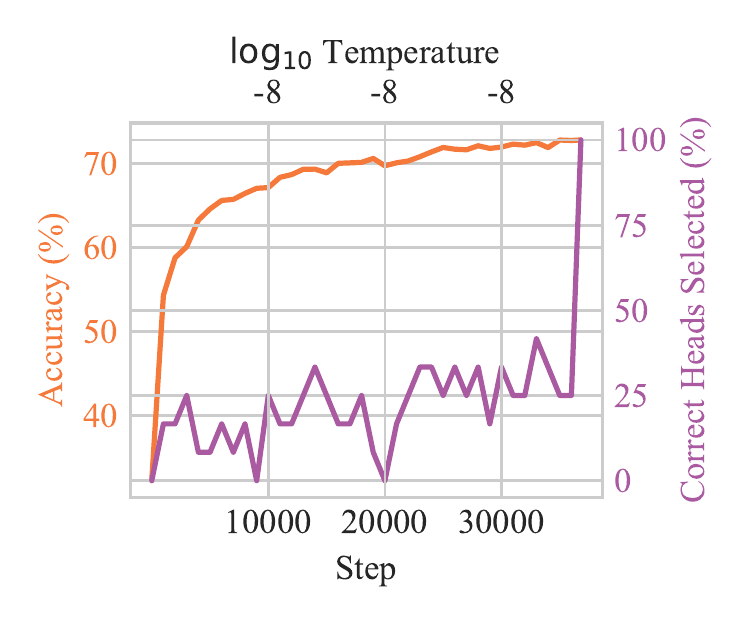}
    \caption{Without Annealing}
    \label{fig:dynammic_no_annealing}
  \end{subfigure}
  \caption{Training dynamics of joint DSP on $\bert$ ($K=12$). The lower $x$-axis shows the number of training steps, and the upper $x$-axis shows the corresponding temperature in logarithm scale. Left $y$-axis (\textcolor{myorange}{orange}) shows test accuracy on MNLI-mismatched validation set. Right $y$-axis (\textcolor{mypurple}{purple}) shows the percentage of heads selected at current step that are kept eventually. }
  \label{fig:dynamic}
  \vspace{10pt}
\end{figure*}

\subsection{Distribution of Heads}
We visualize the distribution of unpruned heads across different layers in \cref{fig:distribution}. For $\bert$ (\cref{fig:distribution_mnli}), we observe that the top layers (10--12) are the first to be pruned and the heads in the middle layers (3--7) are mostly retained. This observation is in conformity with \newcite{prasanna-etal-2020-bert, sajjad2021effect}. \newcite{budhraja-etal-2020-weak} also highlights the importance of middle layers but finds no preference between top and bottom layers. 
For $\ed$ (\cref{fig:distribution_mt}), we find that a lot more encoder--decoder cross attention heads are retained compared to the other two types of attentions (encoder and decoder self attentions). The encoder self-attention heads are completely pruned away when less than $16$ heads are left, which again conforms with the observations of \citet{Michel:2019} and \citet{Voita:2019}.\looseness=-1

\subsection{Analysis of Training Dynamics}

To better understand our joint DSP approach, we inspect its behavior during training. 
We plot the intermediate accuracy 
of $\bert$ during training when joint DSP ($K=12$) is applied %
in \cref{fig:dynamic_annealing} (in \textcolor{myorange}{orange}). 
We also compute the percentage of heads selected
at the current step that are eventually kept in the end (in \textcolor{mypurple}{purple}). 
We observe the selected subset of heads is no longer updated after 14000 training steps (\textcolor{mypurple}{purple line} stays at $100\%$). 
Therefore, the joint pruning process may be viewed as having two distinct phases---i) head selection  and ii) fine-tuning. 
This piques one's interest as it appears to superficially resemble a reversed pipelined pruning.
During head selection, the subset of heads to be kept is determined and the model is adapted to the specified level of sparseness. 
During fine-tuning, the selected subnetwork is fine-tuned so that the testing accuracy improves steadily.
Our experiments indicate that annealing is essential for training a high-performance pruner: It allows the model to gradually settle down on one particular subset of heads, whereas without annealing the pruner never converges to a fixed set
and thereby does not enter the fine-tuning phase. %
See \cref{fig:dynammic_no_annealing} for a visualization.\footnote{We analyze other sparsity levels as well and observe similar behaviors. Two examples are shown in \cref{app:dynamics}.}

\subsection{Summary}
The five pruning methods discussed in this paper are summarized in \cref{table:features}. 
Joint DSP is able to maintain the highest test performance while consuming similar computational resources to \citeauthor{Voita:2019} and offering fine-grained control over the number of unpruned heads like \citeauthor{Michel:2019} It is worth noting that STE shares the same benefits of low computational overhead and exact sparsity control as joint DSP, despite being slightly inferior in performance. It also has less hyperparameters to tune and hence is easier to implement. Therefore, we believe STE could be favorable when test performance is not that critical.

\section{Related Work}

\paragraph{Unstructured Pruning.} Neural network pruning has been studied for decades. Early works include optimal brain damage \cite{lecun} and optimal brain surgeon \cite{Hassibi:93}, which approximate the loss function of a trained model with a second-order Taylor expansion and remove certain parameters in the network while minimizing impact on loss. Recent years have seen a resurgence in this approach \cite{Molchanov:2019, Theis2018Faster, Michel:2019}. 
More recently, magnitude pruning that discards parameters with small absolute values has gained much popularity \cite{han2015learning, han2015deep, guo2016dynamic, zhu2017prune}. \newcite{gordon2020compressing} applies magnitude pruning to BERT and shows that the model has similar prunability and transferability whether pruned after pre-training or after fine-tuning. Related to magnitude based pruning is movement pruning introduced by \newcite{sanh2020movement} which considers changes in weights instead of magnitudes for pruning.

\paragraph{Structured Pruning.} Different from above-mentioned unstructured pruning methods that prune individual parameters, structured pruning methods prune at a higher level, such as convolutional channels, attention heads, or even layers. Structured pruning almost always leads to a decrease in model size and inference cost, while unstructured pruning often results in sparse matrices, which cannot be utilized without dedicated hardware or libraries \cite{han2016eie}. Previously, structured pruning had primarily been applied to convolutional neural networks \cite{wen2016learning, li2017pruning, Luo_2017_ICCV, he2017channel, liu2017learning, huang2018data}, but it has recently been applied to NLP, in the form of layer pruning \cite{fan2019reducing, sajjad2021effect} and head pruning \cite{Michel:2019, Voita:2019, McCarley2019PruningAB} of Transformer-based models. Apart from compression and speedup, head pruning is also helpful for model analysis; \newcite{Voita:2019} finds that the heads that survive pruning play consistent and linguistically-interpretable roles. \newcite{prasanna-etal-2020-bert} discovered the heads that are pruned last tend to be in the earlier and middle layers.

\begin{table*}
\centering\small
 \begin{tabular}{@{}lccc@{}} 
 \toprule
 Methods & Computation Overhead & Sparsity Controllability & Test Performance \\ 
 \midrule
 \citeauthor{Michel:2019} & \faThumbsODown & \faThumbsOUp & \faThumbsODown \\ 
 Pipelined DSP (this paper) & \faThumbsODown & \faThumbsOUp & \faThumbsODown \\
 \citeauthor{Voita:2019} & \faThumbsOUp & \faThumbsODown & \faThumbsOUp \\
 STE (this paper) & \faThumbsOUp & \faThumbsOUp & \faThumbsOUp \\
 Joint DSP (this paper) & \faThumbsOUp & \faThumbsOUp & \faThumbsOUp \\
 \bottomrule
\end{tabular}
\caption{Qualitative Comparison of Different Pruning Methods}
\label{table:features}
\vspace{10pt}
\end{table*}
\paragraph{Dropout for Pruning.} A variety of regularizers have been used to sparsify neural networks. For example, \newcite{han2015learning} applies $L_1$ regularization, and \newcite{louizos2017learning} applies $L_0$ regularization. Dropout, as one of the regularization methods, has also been demonstrated to be effective for converting a model to be robust to pruning. It was discovered that dropout encourages sparsity when dropout was proposed \cite{Srivastava:2014}. Recently, the assumption that the model trained with dropout tend to be more robust to post-hoc pruning was also explored. LayerDrop \cite{fan2019reducing} randomly drops entire layers in Transformer with a fixed dropout rate during training and simply keeps every other layer during inference. Targeted Dropout \cite{Gomez:2019} ranks units in the order of magnitude and only applies dropout to those with small magnitudes and performs magnitude pruning afterwards. \newcite{molchanov2017variational} introduces variational dropout, which allows learning a different dropout rate for each unit. \newcite{Kingma:2015} extends it for pruning by keeping only the units with lower dropout rate for test. Our approach is in the same vein but distinct as we learn importance variables rather than dropout rate and the number of heads to be dropped is specified explicitly, which allows us a control over sparsity.

\paragraph{Lottery Ticket Hypothesis.} \newcite{frankle2018lottery} proposes the Lottery Ticket Hypothesis that there exist subnetworks ("winning lottery tickets") in a over-parameterized model, which can be trained in isolation to reach comparable test performance as the original network in a similar number of iterations. It shows such tickets can be discovered through magnitude pruning. \newcite{brix-etal-2020-successfully} successfully applies the hypothesis to the Transformer. \newcite{prasanna-etal-2020-bert, behnke-heafield-2020-losing} demonstrate head pruning may also be used to select a winning subnetwork.

\section{Conclusion}
We propose differentiable subset pruning, a novel method for sparsifying Transformers.
The method allows the user to directly specify the desired sparsity level, and it achieves a better sparsity--accuracy trade-off compared to previous works, leading to a faster and more efficient model after pruning.
It demonstrates improvements over existing methods for pruning two different models ($\bert$ and $\ed$) %
on two different tasks (textual entailment and machine translation), respectively. 
It can be applied in both pruning paradigms (pipelined and joint pruning).
Although we study head pruning in the paper, our approach can be extended to other structured and unstructured pruning scenarios. In future work, it would be interesting to look into such cases. 

\section*{Acknowledgments}
We would like to thank the action editor Noah Smith and the anonymous reviewers for their helpful comments. MS acknowledges funding by SNF under project \#201009.

\bibliography{tacl2018}
\bibliographystyle{acl_natbib}

\appendix
\section{Experimental Setup}
We report the hyperparameters for joint DSP we use in our experiments in \cref{table:hyper}, which are obtained by tuning on the validation set. 
\label{app:hyper}
\begin{table}[h]
\centering\small
 \begin{tabular}{@{}ccc@{}} 
 \toprule
  & $\bert$ & $\ed$  \\ 
  $\tau_{\text{ini}}$ & $1000$ & $0.1$ \\
  $\tau_{\text{end}}$ & $1e-08$ & $1e-08$ \\
  $N_{\text{cooldown}}$ & $25000$ & $15000$ \\
  lr for $w_h$ & $0.5$ & $0.2$ \\
 \bottomrule
\end{tabular}
\caption{Hyperparameters used for joint DSP.}
\label{table:hyper}
\end{table}

\section{Analysis of Training Dynamics}
\label{app:dynamics}
\begin{figure*}[h]
\centering
  \begin{subfigure}[b]{.49\textwidth}
    \centering\includegraphics[width=\textwidth]{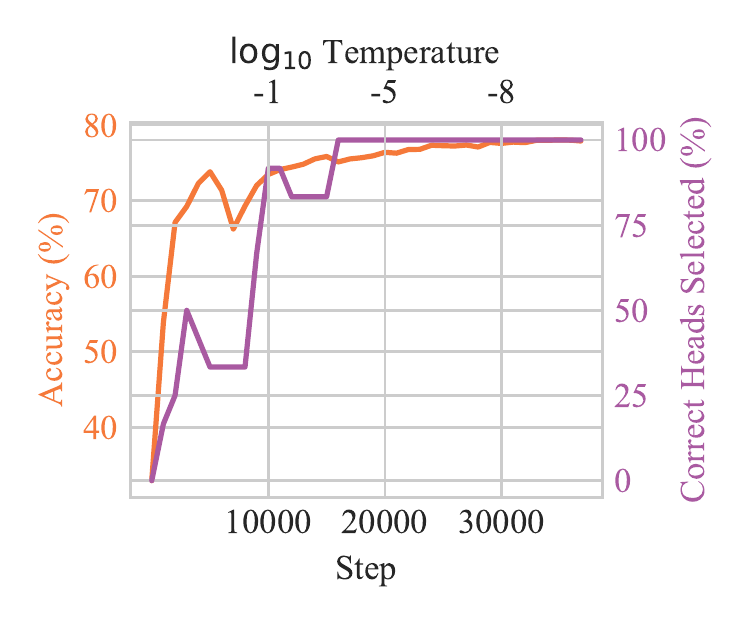}
    \caption{$K=8$ with annealing}
    \label{fig:dynamic_8}
  \end{subfigure}
  \begin{subfigure}[b]{.49\textwidth}
    \centering\includegraphics[width=\textwidth]{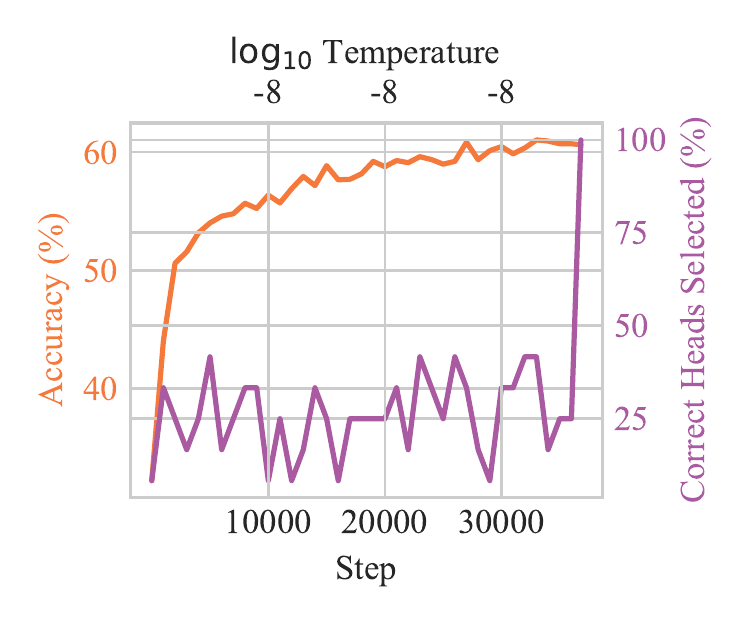}
    \caption{$K=8$ without annealing}
    \label{fig:dynamic_8_woa}
  \end{subfigure}
    \newline
    \vspace{0cm}
    \newline
  \begin{subfigure}[b]{.49\textwidth}
    \centering\includegraphics[width=\textwidth]{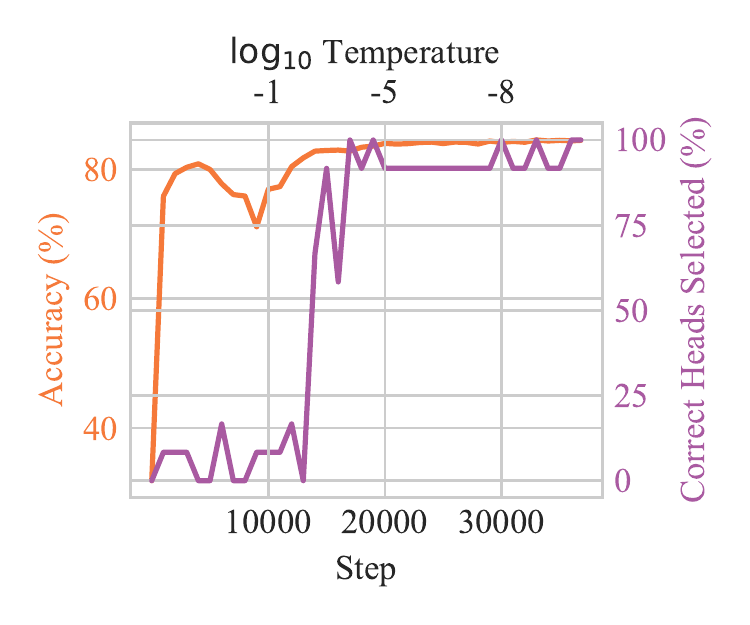}
    \caption{$K=108$ with annealing}
    \label{fig:dynammic_108}
  \end{subfigure}
  \begin{subfigure}[b]{.49\textwidth}
    \centering\includegraphics[width=\textwidth]{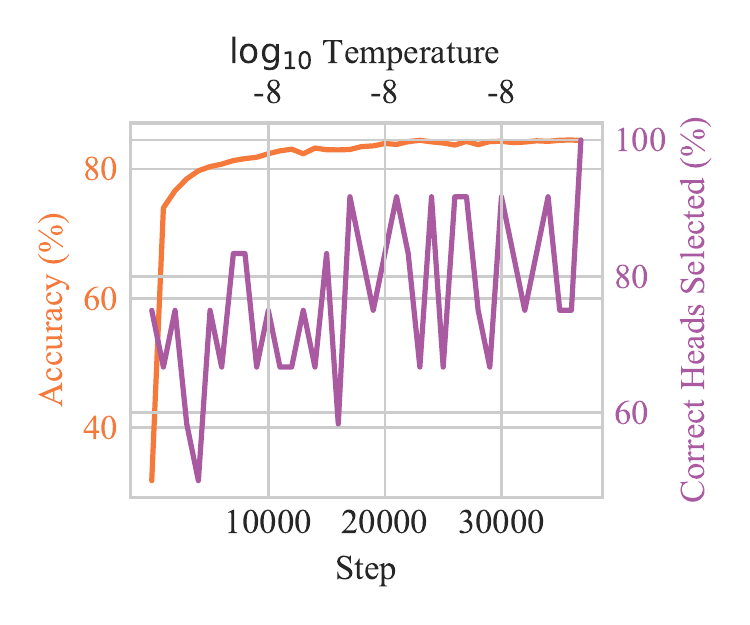}
    \caption{$K=108$ without annealing}
    \label{fig:dynammic_108_woa}
  \end{subfigure}
  \caption{Training dynamics of joint DSP on $\bert$. The lower $x$-axis shows the number of training steps, and the upper $x$-axis shows the corresponding temperature in logarithm scale. Left $y$-axis (\textcolor{myorange}{orange}) shows test accuracy on MNLI-mismatched validation set. Right $y$-axis (\textcolor{mypurple}{purple}) shows the percentage of heads selected at current step that are kept eventually. }
  \label{fig:dynamic_appendix}
\end{figure*}
We present two more examples where heads are scarce ($K=8$) or redundant ($K=108$). In \cref{fig:dynamic_8}, we observe the same two-phase training behavior as $K=12$. The selected subset of heads is not altered anymore after $16000$ steps. In \cref{fig:dynammic_108}, unlike the cases where heads are very few, the head masks are constantly updated throughout the training procedure. Yet a large portion ($91.7\%$) of the heads remain unchanged after $17000$ steps. Its two-phase behavior is still apparent in comparison with training without annealing (\cref{fig:dynammic_108_woa}).

\section{Detailed Results}
\label{app:results}
The detailed results for plotting \cref{fig:comparison} are presented in \cref{table:comparison}. 
\begin{table*}
\begin{subtable}[h]{\textwidth}
\centering\small
 \begin{tabular}{@{}cccccc@{}} 
 \toprule
 Unpruned Heads & \citeauthor{Michel:2019} & Pipelined DSP & \citeauthor{Voita:2019} & STE & Joint DSP \\ 
 \midrule
    132 & 84.38 & 84.15 & 84.26 & \textbf{ 84.77 } & 84.70 \\
    120 & 84.60 & 84.41 & 84.18 & 84.59 & \textbf{ 84.97 } \\
    108 & 84.19 & 82.64 & 84.39 & \textbf{ 84.52 } & 83.95 \\
    96 & 84.24 & 83.27 & 84.42 & \textbf{ 84.68 } & 84.41 \\
    84 & 83.50 & 83.37 & 84.00 & \textbf{ 84.20 } & 84.02 \\
    72 & 82.47 & 82.95 & 83.93 & \textbf{ 84.08 } & 83.48 \\
    60 & 81.74 & 79.69 & 83.37 & \textbf{ 83.85 } & 83.21 \\
    48 & 79.26 & 79.10 & \textbf{ 83.24 } & 82.81 & 83.22 \\
    36 & 70.82 & 76.08 & 81.68 & 82.20 & \textbf{ 82.51 } \\
    24 & 47.54 & 70.72 & 81.02 & 81.44 & \textbf{ 81.54 } \\
    12 & 40.59 & 56.29 & 76.91 & 73.79 & \textbf{ 79.74 } \\
    11 & 40.16 & 50.81 & 76.30 & 78.91 & \textbf{ 79.02 } \\
    10 & 39.71 & 49.14 & 75.34 & 77.10 & \textbf{ 78.35 } \\
    9 & 40.88 & 51.20 & 76.12 & 76.99 & \textbf{ 77.51 } \\
    8 & 36.16 & 45.74 & 74.12 & 69.29 & \textbf{ 77.57 } \\
    7 & 36.13 & 43.11 & 74.14 & 69.64 & \textbf{ 76.32 } \\
    6 & 34.28 & 40.90 & 74.18 & 70.45 & \textbf{ 76.70 } \\
    5 & 33.24 & 41.95 & 73.89 & 66.53 & \textbf{ 76.17 } \\
    4 & 33.49 & 42.64 & 73.12 & 65.43 & \textbf{ 75.06 } \\
    3 & 32.68 & 41.79 & 62.84 & 65.15 & \textbf{ 73.36 } \\
    2 & 32.74 & 38.30 & 62.87 & 57.07 & \textbf{ 72.14 } \\
    1 & 34.28 & 43.28 & \textbf{ 62.09 } & 61.79 & 61.79 \\
 \bottomrule
\end{tabular}
\caption{Accuracy on the MNLI-mismatched validation set as a function of number of remaining heads in $\bert$.}
\label{table:mnli}
\end{subtable}
\newline
\vspace{1cm}
\newline
\begin{subtable}[h]{\textwidth}
\centering\small
 \begin{tabular}{@{}cccccc@{}} 
 \toprule
 Unpruned Heads & \citeauthor{Michel:2019} & Pipelined DSP & \citeauthor{Voita:2019} & STE & Joint DSP \\ 
 \midrule
    68 & 32.87 & 34.19 & 34.10 & \textbf{ 34.69 } & 34.52 \\
    64 & 29.08 & 34.29 & 34.19 & \textbf{ 34.55 } & 34.51 \\
    60 & 11.18 & 32.21 & 34.14 & 34.56 & \textbf{ 34.83 } \\
    56 & 6.91 & 32.52 & 34.19 & 34.19 & \textbf{ 34.46 } \\
    52 & 4.41 & 33.02 & 34.23 & 33.92 & \textbf{ 34.79 } \\
    48 & 2.64 & 31.58 & 34.20 & 34.02 & \textbf{ 34.82 } \\
    44 & 2.30 & 28.70 & 34.08 & 33.88 & \textbf{ 34.68 } \\
    40 & 1.70 & 24.35 & 34.06 & 33.85 & \textbf{ 34.13 } \\
    36 & 1.20 & 25.84 & 33.82 & 33.22 & \textbf{ 34.58 } \\
    32 & 0.61 & 23.94 & 33.70 & 32.88 & \textbf{ 34.10 } \\
    28 & 0.19 & 16.63 & 33.78 & 32.01 & \textbf{ 33.89 } \\
    24 & 0.13 & 20.40 & 33.44 & 33.71 & \textbf{ 33.72 } \\
    20 & 0.07 & 14.11 & 33.25 & 31.27 & \textbf{ 33.54 } \\
    16 & 0.07 & 7.55 & \textbf{ 32.62 } & 31.25 & 32.32 \\
    12 & 0.05 & 3.80 & 32.33 & 30.71 & \textbf{ 32.74 } \\
    8 & 0.04 & 0.63 & 31.26 & 28.77 & \textbf{ 32.68 } \\
    4 & 0.04 & 0.16 & 29.09 & 25.45 & \textbf{ 30.33 } \\
    3 & 0.04 & 0.09 & 23.08 & 23.83 & \textbf{ 28.22 } \\
    2 & 0.04 & 0.05 & 20.89 & 22.35 & \textbf{ 24.18 } \\
    1 & 0.04 & 0.05 & 20.38 & 20.37 & \textbf{ 20.64 } \\
 \bottomrule
\end{tabular}
\caption{BLEU score on IWSLT test set as a function of number of unpruned heads in $\ed$.}
\label{table:mt}
\end{subtable}
\caption{A comparison of various pruning methods.}
\label{table:comparison}
\end{table*}

\end{document}